\documentclass[A Survey for Hierarchical Multi-label Text Classification]{llncs}
\usepackage[dvipsnames]{xcolor}
\usepackage{graphicx}

\begin{document}
\title{Recent Advances in Hierarchical Multi-label Text Classification: A Survey}
%
%\titlerunning{Abbreviated paper title}
% If the paper title is too long for the running head, you can set
% an abbreviated paper title here
%
%\author{Anonymous Author}
\author{Rundong Liu \and
Wenhan Liang \and Weijun Luo\and Yuxiang Song \and He Zhang \and Ruohua Xu \and Yunfeng Li \and Ming Liu}

\authorrunning{Rundong Liu, et al.}
% First names are abbreviated in the running head.
% If there are more than two authors, 'et al.' is used.
%

\institute{Beijing University Of Posts and Telecommunications, Beijing, China \\
Patent Examination Cooperation, CNIPA, Henan, China \\
Zhongtukexin Co. Ltd., Beijing, China \\
Deakin University, Melbourne, Australia \\
\email{songyuxiang@cnipa.gov.cn \\
\{zhanghe, xuruohua, liyunfeng\}@kxsz.net \\
m.liu@deakin.edu.au\\liangwh2013@163.com\\overtheriver861@gmail.com\\liurundong991@bupt.edu.cn}\\
}

\maketitle              % typeset the header of the contribution

\begin{abstract}

Hierarchical multi-label text classification aims to classify the input text into multiple labels, among which the labels are structured and hierarchical. It is a vital task in many real world applications, e.g. scientific literature archiving. In this paper, we survey the recent progress of hierarchical multi-label text classification, including the open sourced data sets, the main methods, evaluation metrics, learning strategies and the current challenges. A few future research directions are also listed for community to further improve this field. 

\end{abstract}

\section{Introduction}
The recent advances in deep learning, large language modeling, instruction learning and human-AI alignment has revolutionized the NLP community. Among all these different kinds of NLP tasks, text classification is still regarded as the backbone for most downstream NLP applications. In the real world scenario, most text classification tasks are multi-label ones. For example, the hierarchical categorization of scientific literature in arXiv, the grouping of products in online shopping sites, as well as the management of papers in e-libraries. Conceptually,  hierarchical multi-label text classification means that text are associated with multiple labels, for which the labels are organized in a structured hierarchy. Mathmatically, hierarchical multi-label classification can be defined as a function: $ f: X \rightarrow Y^*$,  where $X$ is the sample space and $Y^*$ is the power set of all possible label sets. Each element in $Y^*$ is an ordered tag sequence $(y_1, y_2,..., y_k)$, representing the hierarchical path from the root node to the leaf node. The goal of hierarchical multi-label classification is to learn a function $f$, so that for any given sample $x \in X$, $f (x)$ can output one or more labels which are most relevant to $x$. Traditionally, the one-vs-rest or one-vs-one learning strategy is a common way to learn hierarchical multi-label classifiers, but it ignores the information of the label hierarchy and the learning process is expensive. Modern approaches utilize a deep learning encoder based on a Binary Cross Entropy loss and do the training in a end-to-end manner. In addition, the label information can also be leveraged in the learning process. 

Currently, there are more advanced models and learning approaches for hierarchical multi label tasks, but few literature has summarized these methods. Therefore, we conduct a survey of recent advances on hierarchical multi-label text classification. This survey is organized in the following way: Section 2 summarizes the data sets used for hierarchical multi-label classification. Section 3 mainly discusses the models and methods used for hierarchical multi-label text classification, which are mainly divided into four major categories. In section 4, we discuss some machine learning strategies for hierarchical multi-label text classification. Subsequently, section 5 show the widely used evaluation metrics and section 6 will mention some existing challenges and the last section is the conclusion and some suggestion  for future work.
\section{Hierarchical Multi-label Classification Data sets}
In this section, we will introduce the common hierarchical multi-label classification data sets. In total, 13 data sets are included and Table 1 shows the text domain, number of labels and label depth of each hierarchical multi-label data set. 
\begin{table}
\centering
\caption{A summary of hierarchical multi-label data sets}
\label{tab1}
\begin{tabular}{|c|c|c|c|}
\hline
Dataset &  Domain & Number of labels & Label depth \\
\hline
YELP & dining reviews & 539 & -\\
%Amazon 670k & product reviews & 670,091 & 9\\
Amazon 3M & product reviews & 2,812,281 & - \\
IMBD-Multi & movie reviews & 81,742 & 21\\
WOS  & academic articles & - & -\\
% WOS-11967 & academic articles & 40 &2\\
Arxiv & academic issues & - & -\\
Pubmed & biomedical articles & 17693 & 15\\
Reuters  & news & - & -\\
%RCV1 &Reuters news &103 &6\\
NYTimes & news & 2318 & 10\\
HUPC  & patent & - & - \\
BGC-EN  & books & 146 & 4 \\
WIPO-alpha & patent &5229 &4 \\
DBpedia & wikipedia  & - & - \\
SciHTC & scientific articles & 1233 & -\\
%Delicious & users  & 983 & - \\
%Enron & email & 56 & 3\\
\hline
\end{tabular}
\end{table}

The YELP data set \cite{10.1145/3439726} is a famous sentimental data set,which sourced from the US restaurant review website. User reviews of YELP-2 mainly use positive and negative emotion labels, while YELP-5 reviews have five different scores. Amazon Product Reviews \cite {NEURIPS2021_3bbca1d2} contains a large number of user product reviews of Amazon, with about 8 million reviews. IMDB-Multi \cite{10.1145/3439726} has the meta data for 81,742 movies, each movie can have different genres. 
The SciHTC\footnote{https://arxiv.org/pdf/2211.02810v1.pdf}  contains 186,160 papers and 1,233 categories from the ACM CCS tree. The WOS data set\footnote{https://arxiv.org/pdf/1709.08267v1.pdf} contains web of science articles, with 134 catalogues and 46,985 instances, covering all aspects. Pubmed\footnote{https://pubmed.ncbi.nlm.nih.gov/download/} is from the biomedical domain and  researcher can retrieve the latest literature from the official NIH website . HUPC\footnote{https://arxiv.org/pdf/2207.04043.pdf} and WIPO\footnote{https://www.wipo.int/classifications/ipc/en/ITsupport/Categorization/dataset/index.html} are patent data sets,including patents'ID,authors,classes.
Reuters data set has two versions: RCV1\footnote{https://dl.acm.org/doi/pdf/10.5555/1005332.1005345} is modified on the Reuters-21578 data set and RCV2 has nearly 5 million news in 13 different languages. NewYork Times data set\footnote{https://catalog.ldc.upenn.edu/LDC2008T19} contains 1.8 million articles from New York Times during 1987-2007.
Blurb Genre Collection (BGC)\footnote{https://www.inf.uni-hamburg.de/en/inst/ab/lt/resources/data/ blurb- genre- collection.html} is a data set of book advertising descriptions and it consists of 91,892 instances, including the book’s title, the author, url, ISBN, etc.DBpedia\footnote{https://www.researchgate.net/publication/259828897\_DBpedia\_-\_A\_Large-scale\_Multilingual\_Knowledge\_Base\_Extracted\_from\_Wikipedia} is extracted from Wikipedia. It is updated monthly and the label scheme is adjusted as well.
\section{Main Approaches for hierarchical multi-label classification}
In this section, we consider four main approaches for hierarchical multi-label text classification: tree-based approach, embedding based approach, graph based approach and the ensemble approach.        
\subsection{Tree-based approach} This method is based on probabilistic label tree, which was originally developed for extreme multi-label classification, where a probabilistic label tree (PLT) is used to partition labels, where each leaf in PLT corresponds to an original label and each internal node corresponds to a pseudo-label (meta-label). Then by maximizing a lower bound approximation of the log likelihood, each linear binary classifier for a tree node can be trained independently with only a small number of relevant samples. Parabel \cite{prabhu2018parabel} is a traditional label tree-based method using bag-of-words (BOW) features. FASTTEXT \cite{joulin2016bag} and EXTREMETEXT\cite{wydmuch2018no} extended Parabel by using dense features.  AttentionXML \cite{you2019attentionxml}, which used attention model and a PLT to further improve the classification performance. FastXML \cite{prabhu2014fastxml} draws on thee multi-label random forest (MLRF) algorithm and sublinear ranking label (LPSR) partition algorithm. Bonsai \cite{2019arXiv190408249K} proposed to build a shallow and diverse label tree structure, in which all the labels are segmented into K sets through the clustering, but it needs high space complexity because of using linear classifiers.
HPT \cite{wang-etal-2022-hpt} make efforts to transform hierarchy text classfication(HTC) into a hierarchy-aware multi-label mask language model(MLM) problem that focuses on bridging two gaps between HTC and MLM.
%In order to make good use of label hierarchy information, HPT not only constructed a template based on the depth of label hierarchy as input, but also utilized a K-layer Graph Attention Network(GAT) model to adopt relationships between labels.}

For the tree-based methods, their advantages are that they can use the hierarchical structure between labels more accurately and the prediction results of the model are easy to explain. The disadvantage is that there is a high requirement for label level knowledge that needs to be acquired in advance and accurate hierarchy information is required. Besides, because it uses a relatively simple tree structure model, it is difficult to deal with the complex dependency relationship between labels, which needs to be represented by a more powerful model.
\subsection{Embedding based approach}
Embedding based methods arise with deep learning models, most of these methods, use an encoder to represent the input text and learn a mapping function for input text and the labels. HyperIM \cite{DBLP:journals/corr/abs-1905-10802} embeds both document words and labels jointly in the hyperbolic space to preserve their latent structures. The innovation of this method is that it can connect words in documents and labels, and use the most relevant part of documents to build the representations of the article, so as to improve the accuracy of classification. HTrans \cite{banerjee-etal-2019-hierarchical}, a hierarchical transfer learning approach uses a Gated Recurrent Unit based  architecture coupled with attention mechanism. In Hierarchical Fine-Tuning (HFT) \cite{shimura-etal-2018-hft} the parameters of the word embedding layer and the convolution layer are mainly transferred from top to bottom, that is, the parameters of the upper layer structure are used to learn the labels of the lower layer. At the same time, in order to avoid adverse effects caused by excessive correlation gaps between distant levels, parameter fine-tuning is only performed between adjacent levels. C-HMCNN \cite{2020arXiv201010151G} uses a constraint loss function to make the hierarchical structure of the lower layer to predict the upper layer. LA-HCN \cite{2020arXiv200910938Z} uses the label-based attention to bridge the labels and document content. Unlike the features extracted from typical attention may be diluted, it can extract meaningful information corresponding to different labels to learn learning disjoint features for each hierarchy. The HMCN \cite{wehrmann2018hierarchical} simultaneously optimizes local and global loss functions for discovering local hierarchical class-relationships and global information from the entire class hierarchy while penalizing hierarchical violations.

The embedding based approach often needs a large amount of training examples, it relies on the attention mechanism and regularization from the label structure. A weakness is that there is no explainability for wrongly predicted labels.  
\subsection{Graph based approach}
Graph based approaches regard the whole document and hierarchical labels as a single graph, where the node type may include token, label, sentence, topic or even some meta information. A graph neural network is then applied on the graph and the hierarchical multi-label classification task can be converted into a node classification task. For example, heterogeneous graphs are used in \cite{ye-etal-2021-beyond}  to model the labels' hierarchy, their statistical dependencies and meta data.  Loosely coupled graph convolutional neural network (LCGCN) \cite{xu-etal-2021-hierarchical} mainly solves the over-smoothing problem caused by too many nodes or edges of the GCN graph model. There are two parts in it, one is the core graph, which is used to extract word embeddings and label embeddings, the other one is the document-word graph, which is mainly used to capture the relationship between documents and words. In another hierarchy-aware global model \cite{zhou-etal-2020-hierarchy}, the label hierarchy is represented as a directed graph with two sub models: HiAGM-LA (the label  features can be fused with each other, and the text and label are independently expressed and predicted by attention mechanism) and HiAGM-TP (text features can be directly input into the structural encoder to participate in the feature calculation). This model captures the hierarchical information and fully utilizes the interaction of text features and label features.  

The graph based approach inputs the hierarchical information of labels into the network as known knowledge, and some documents also use the meta data of documents as input. The advantage of this method is that it can better handle the complex dependency between tags. Besides, the model design structure is not complex, which only needs to model and represent the hierarchical structure. However, the classifier cannot make more targeted use of the hierarchical structure between labels. Similar with the embedding approach, the prediction results of the model are difficult to explain.
\subsection{Ensemble approach}
Ensemble is widely used in statistical machine learning. In terms of hierarchical multi-label classification, several models can be trained for each of the levels of the labels, which can be conducted in parallel. In the prediction time, a voting mechanism can be conducted, followed by further cross checking among the predicted labels. However, this method is mostly used in the industrial field and is suitable for large engineering projects.

Among different ensemble approaches, one-vs-one \cite{daengduang2017applying} and one-vs-all \cite{10.1007/978-3-031-17189-5_14} are the two typical ways. One-vs-one refers to learning a classifier between each two categories. It works well when the total amount of labels is small, but as the number of levels of labels increases, the number of classifiers also increases dramatically. When there are n categories here, then the number of the classifiers will grow to $C_{n}^{2}$. In contrast, one-vs-all means that a single classifier is trained for each label, and the classifier treats its corresponding category as positive examples and treats the remaining categories as negative examples. In the shared task 5 track 1 of NLPCC2022 multi-label classification for scientific literature \cite{liu2022overview}, the task requires participants to build multi-label text classification models for scientific abstracts from the chemistry domain. Among all the submitted system, the best system \cite{wang2022bit} was obtained by an ensemble model with more than five different deep learning models. Another boosting ensemble method is derived from Devil's advocate \cite{jo-etal-2021-devils-advocate}. This method requires at least three models to be integrated, and a new loss which uses the idea of generated confrontation to make the target model finally converge is designed.  
%The paper\cite{10.1007/978-3-031-17189-5_14} uses this method. The Author proposed two modules: a shared encoding module and a task-specific module. First, the language model SciBERT was pre-trained. The vector obtained by encoding the text feed into the task-specific module, where the sub-task of classifying each tag is performed. And then a shared TextCNN layer was introduced to learn the dependency information between labels at each level. Finally, the hierarchical feature information is fused by linear transformation and the prediction result of each level is getted.
\section{Machine Learning Strategies}
% Who write this seciton, very good. 
Many machine learning strategies were adopted for hierarchical multi-label classification. These learning strategies usually consider multiple aspects, such as data, label and training. We list the following machine learning strategies. 
\paragraph{\textbf{Self-training}} Self training is a semi supervised learning method in which a small amount of labeled data is used for initial training, and then prediction results are used to generate more training data. In hierarchical multi label classification tasks, self training can generate more training data by using the prediction results of the current model, and then use these data for the next round of training.
%\subsection{Semi-supervised} Semi-supervised learning is a method of using unmarked data to enhance supervised learning. In hierarchical multi label classification tasks, semi supervised learning can use unlabeled data to learn hierarchical relationships between labels, and then use these relationships in supervised learning.\\

\paragraph{\textbf{Weakly-supervised learning}} In reality, a real situation is that there is no big amount of human labeled documents to train classifiers, so approaches based on only class surface names as supervision signals are explored. An example is TaxoClass \cite{shen-etal-2021-taxoclass}.  First, a pre-trained textual entailment model is used to  calculate the document-class similarity. After that, the calculated similarity is used to determine the core class of the article. Then training data extracted from document core classes is used to train a text classifier. The classifier includes a document encoder based on pre-trained BERT, a class encoder capturing class taxonomy structure, and a text matching network computing the probability of a document being tagged with each class. The class Encoder is a graph neural network to capture the hierarchical information of label. This method can train a large number of unlabeled documents, but because of the unlabeled data, the determination of the core class of the document may not be so accurate, which may affect the final accuracy. ASR2 \cite{song-etal-2022-adaptive} is proposed as a general, model-agnostic weakly supervised leading framework, which is dedicated to alleviate the data imbalance issue. It calculates a probabilistic margin score based on the output of the current model to measure and rank the cleanliness of each data point. Then, the ranked data is sampled based on both class-wise and rule-aware ranking. Another example \cite{Meng_2018} is the use of constructed pseudo-documents to address label sparsity by improving the dataset in three different ways. Assuming that words and documents are located in the same semantic space, a spherical distribution is established to sample keywords that generate pseudo-documents, thereby improving the generalization of seed information.

\paragraph{\textbf{Data augmentation}}
Data augmentation is a commonly used training strategy that can increase the diversity of training data and help models better generalize to new data. In hierarchical multi label classification tasks, data augmentation can be used to generate more high-quality training data, and can help models learn hierarchical relationships between labels by generating data related to hierarchical relationships. Data enhancement can be implemented in the following ways: Synonym Replacement, Random Insertion, Random Deletion, Random Swap, and Sentence Rearrangement. The above methods can be randomly applied to training data with a certain probability, thereby expanding the data set and increasing the generalization ability and robustness of the model. At the same time, it is also important to note that excessive data enhancement may lead to over fitting of the model, so it is necessary to make choices based on specific tasks and data characteristics.

\paragraph{\textbf{Label enhancement}}
Label enhancement is a method of enhancing original label information through the use of external resources (such as knowledge maps, dictionaries, corpora, etc.). In hierarchical multi label classification tasks, label enhancement can help models more accurately understand and depict the relationship and semantic information between different labels, thereby improving the classification accuracy and generalization ability of the model. However, it should be noted that excessive label enhancement may introduce noise and incorrect labels, thereby reducing the performance and generalization ability of the model. Therefore, this method needs to be used appropriately. MHG \cite{ye-etal-2021-beyond}
modeles metadata and its topological relationships to construct a metadata heterogeneous graph,to capture label dependencies. HVHMC \cite{xu-etal-2021-hierarchical} integrates hierarchical information through graph convolutional network learning as a way to capture the horizontal and vertical dependencies between labels.
\paragraph{\textbf{Post-processing}}
In hierarchical multi-label text classification, post processing refers to a strategy of adjusting the relationship between labels when the model has already predicted multiple labels. It is usually used to solve situations where there is inconsistency or conflict between the labels predicted by the model. Common post processing strategies include label filtering (remove predicted untrustworthy or irrelevant labels by setting thresholds or other rules), label combination (combine the predicted labels into higher-level labels to reflect their relationship), label conflict (post-processing predicted labels to resolve conflicts or inconsistencies between them). The post-processing steps may vary according to the hierarchy of the label scheme. 
% 所有段落都将取消首行缩进
\setlength{\parindent}{0pt} %
% 设置每个段落之间的间距
\setlength{\parskip}{1em}

\section{Evaluation Metrics for Hierarchical Multi-label Classification.}
For a hierarchical multi label classification task, three types of metrics are often used: \textbf{label-based}, \textbf{example-based} and \textbf{ranking based}. Precision, recall, and F1-score are often used as label-based evaluation metrics. Example-based evaluation metrics includes \textit{accuracy} , \textit{hamming loss} , \textit{subset accuracy} and \textit{information contrast model} \cite{amigo-delgado-2022-evaluating}. Ranking-based evaluation metrics includes \textit{Precision@k} and \textit{Mean Reciprocal Rank}. \textit{Precision@k} and \textit{mean reciprocal rank} are two widely used evaluation metrics for assessing the performance of machine learning classifiers and ranking tasks, respectively. 

\section{Challenges}
Even though many methods have been proposed, there are still many challenges for hierarchical multi-label text classification, including but not limited to the label sparsity and imbalance, low resources labeling data, low accuracy in deep levels of labels, and the extreme multi-label problem. 

\textbf{Label sparsity and imbalance} Label sparsity and imbalance refers to the long-tail distribution of the labels, by which only a small set of labels have many training examples and a large number of the labels are low frequent ones. The model can easily over-fit the high-frequency labels and under-fit the low frequency ones. Often times, when a label appears only a few times, the classifier may ignore it and the general performance is not affected. The model will be more reliable to the data sets with relatively balanced distribution of labels. Some models try to alleviate the imbalance of labels at this level by utilizing predicted labels from the previous level or leveraging global information, such as HMCN-F \cite{inproceedings}, HTF-CNN \cite{shimura-etal-2018-hft}. However, due to the use of a large number of parameters and the lack of overall information, these methods are prone to exposing bias issues. HiAGM \cite{zhou-etal-2020-hierarchy}, improved with a bidirectional computing structure encoder, enhances its ability to handle sparse data. HiLAP \cite{Mao_2019} adopts reinforcement learning to to learn a Label Assignment Policy. ARS2 \cite{song-etal-2022-adaptive} is a framework which proposed two adaptive sampling strategies to address data imbalance issues. Meta-LMTC \cite{wang-etal-2021-meta-lmtc} is used to solve scenes with few and zero shots. Using a meta-learning approach with fine-tuning of parameters to quickly make labels with fewer instances adaptable to the task. And the balancing loss functions for multi-label text classification \cite{huang-etal-2021-balancing} is also introduced to deal with long-tailed distributions.
%\textbf{Weak correlation among the labels}Label weak correlation means the weak correlation between labels at different levels or within the same level, that is, the dependency between labels is not obvious or difficult to capture. This weak correlation is a hidden information but important for the classification accuracy.

\textbf{Low resource labeled data} The lack of labeled data another challenge. Zero-shot learning is a common setting, which refers that some labels do not have corresponding training data, only descriptions of the labels.  A normal method is to convert it into a nearest neighbor search problem. Currently, one model \cite{chalkidis-etal-2020-empirical} based on LWAN by using the label hierarchy is aimed to improve zero-shot learning. An end-to-end structural contrasting representation learning approach \cite{zhang-etal-2022-structural-contrastive} is proposed by raising a novel randomized text segmentation method. However, the enormous label space and complex association between labels and text lead to extensive calculations and low accuracy. 

\textbf{Lower accuracy in deeper levels} In most hierarchical multi-label classification tasks, deeper level labels involve more specific concepts or categories, and these categories may appear less frequently in the data set, which makes it difficult for the model to extract deep features. Besides, the incorrect prediction at one level may affect the results of its child labels. The error prediction can be propagated to the deeper levels, which causes low accuracy in deep level label prediction. 
%\subsection{Computational performance}
%Because a large number of complex network models are used, such as convolution neural network, recurrent neural network, graph structure and tree structure, there are a lot of parameters to learn, which causes much time and resources on training.Thus a more important thing is to simplify model. 

\textbf{Extreme multiple label problem}
The extreme multi-label text classification is the problem of labeling an instance with a small subset of relevant labels chosen from an extremely large pool of possible labels. Due to the lack of enough training data, clustering and ranking algorithms are often leveraged before the classification step, such as Divide and Conquer \cite{barros-etal-2022-divide} and Seq2Set \cite{cao-zhang-2022-otseq2set}.  

%\textcolor{blue}{A recent method named Divide and Conquer\cite{barros-etal-2022-divide} using Matcher module to allocate the probability of documents belonging to each cluster which created by using hierarchical information. Then the Ranker module assign labels only using the documents in the cluster. }\textcolor{blue}{Seq2Set\cite{cao-zhang-2022-otseq2set} uses the bipartite matching and the optimal transport distance ,through an attention mechanism to generate label sets autoregressively, solving the problem that fully connected layers can not generate variable-length label sets.}
\section{Conclusions and Future Work}
In this paper, we conduct a survey for hierarchical multi-label text classification, including the data sets, main approaches, learning strategies, evaluation metrics and challenges.  We find that early hierarchical multi label classification models were based on a tree structure, and more recent approaches rely on deep learning models, especially Transformer and BERT.  Currently, the graph neural network based approach is the main stream, where the hierarchical label information is modeled together with the input text in a graph and a deep graph neural network is learned. The current state of art uses an ensemble approach, where a set of classifiers based on the hierarchical structure of labels are learned, and a global classifier is then developed to predict all the labels. The ensemble approach is quite effective but few formal results are published in academia. We consider a few future research directions: i) Leverage instruction learning for hierarchical multi-label text classification, as prompt learning has shown quite strong performance. Designing appropriate prompt for hierarchical multi-label classifiers can be regarded as a weakly supervised learning problem.  ii) Deal with the label imbalance and sparsity problem.  iii) Inject domain knowledge into the classifier, this is often efficient when labeled data is scare and domain knowledge is strait-forward.  iv) As recent big language models like ChatGPT and GPT4 are released, the ability of zero-shot learning of these language models remains as question. v) Incremental hierarchical multi-label learning, the label hierarchy can evolve with more labels and different hierarchy, continual learning of the classifier is a key problem in the real world scenario. 

%\\In the future, how to design a good classifier is a research direction. A good classifier can make good use of the relationship between text and layers of labels, as well as the dependency relationship between layers of labels. Secondly, finding a more suitable text encoding method can be regarded as a research direction, and whether there is a better text encoding method that can enable the network to better understand the information within the text. At the same time, as research continues to deepen, the problem of extreme multi label classification will become increasingly prominent, and the problem of long tailed data distribution will become increasingly prominent. Therefore, how to solve these two issues is also one of the future research directions.

%
% ---- Bibliography ----
%
% BibTeX users should specify bibliography style 'splncs04'.
% References will then be sorted and formatted in the correct style.
%
% \bibliographystyle{splncs04}
\bibliographystyle{plain}
\bibliography{mybib.bib}

\end{document}